2014 World Congress on Computing and Communication Technologies# Sparse Coding: A Deep Learning using Unlabeled Data for High - Level Representation

Mrs. R. Vidya M.Sc., M.Phil,
Asst. Prof / PG and Research Department of Computer science
St. Joseph's college of Arts and Science (Autonomous)
Cuddalore-1&
Research scholar in MS University, Tirunelveli
vidya.sjc@gmail.com

Dr.G.M.Nasira
Asst.Prof/Department of Computer Applications/Chikana Govt. Arts College.
*Tirupur*
nasiragm99@yahoo.com

Ms. R. P. Jaia Priyankka
M.Phil Computer Science Scholar
St. Joseph's college of Arts and Science
(Autonomous)
Cuddalore-1
jaiapriyankka@yahoo.com*Abstract*- Sparse coding algorithm is an learning algorithm mainly for unsupervised feature for finding succinct, a little above high - level Representation of inputs, and it has successfully given a way for Deep learning. Our objective is to use High - Level Representation data in form of unlabeled category to help unsupervised learning task. when compared with labeled data, unlabeled data is easier to acquire because, unlike labeled data it does not follow some particular class labels. This really makes the Deep learning wider and applicable to practical problems and learning. The main problem with sparse coding is it uses Quadratic loss function and Gaussian noise mode. So, its performs is very poor when binary or integer value or other Non-Gaussian type data is applied. Thus first we propose an algorithm for solving the L1 - regularized convex optimization algorithm for the problem to allow High - Level Representation of unlabeled data. Through this we derive a optimal solution for describing an approach to Deep learning algorithm by using sparse code.

*Keywords: Deep Learning, High - Level Representation, Sparse Coding, Unlabeled Data, Neural Network.*## I. INTRODUCTION

Deep Learning problem arises, when we have limited labeled data for classification task and also large amount of unlabeled data available which is very mildly related to the task. Like labeled data, unlabeled data does not share same class label or some distribution but with specific Sparse coding we can convert this mildly related unlabeled data to perform some classification task[1]. Sparse coding model input consist of real valued vector and it can be described well using Gaussian model and this Gaussian's Sparse coding is really best one among it. When Sparse coding is subjected to unlabeled data it comes to be a great problem and it turns into non-Gaussian form[2].

Our parameter makes the learning problem significantly harder. However this problem can be solved by an optimization process of $L_1$ - Regularized Convex Optimization algorithm, which will solve this non-Gaussian error and this optimization procedure can be applied to other $L_1$ Regularized Optimization problem and it is especially efficient for the problems which have very sparse optimal solution

We successfully apply this model to Deep learning problems with unlabeled data for High - Level Representation. labeled data for machine learning is very difficult and expensive to obtain but the ability to use unlabeled data makes a significance growth in the terms of learning method[3]. To motivate our discussion let's consider some example of "Giraffe".

In Fig 1,when certain pictures of giraffe were used it is easy to find out labeled data whereas other type of unlabeled data remains to be a invariant recognition in natural images. In the image it is not easy to identify giraffe, because it is not in the form of reality[6]. we pose Deep learning problem mainly to machine learning framework which will then have the capacity to make learning easier. Our approach uses the unlabeled data to learn high level representation.

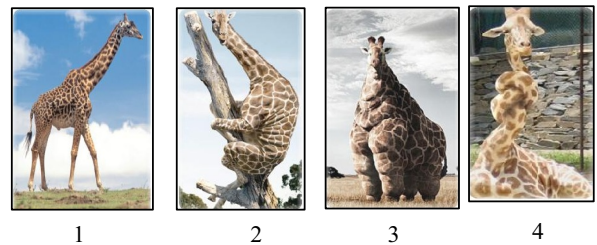

Fig 1. 1st Image is a unlabeled image which can be identified during normal search where as other unlabeled images cannot be identified.

## II. LEARNING : UNSUPERVISED & SUPERVISED

Unsupervised Learning under machine learning refers to be a problem of finding the structure in the unlabeled data which is hidden. The paradigm given to the learners

124978-1-4799-2877-4/14 $31.00 © 2014 IEEE
DOI 10.1109/WCCCT.2014.69

are in the form of unlabeled data, so there are no possible ways to find out the potential solution[10]. This totally distinguishes the supervised learning from unsupervised learning. Unsupervised method is intimately linked to the inference of thickness or density in statistics. In unsupervised methodology there are variety of techniques which summarize the main features in data.

### A. Supervised Vs Unsupervised

Theoretical point of view says that Supervised and Unsupervised learning differs mainly from their model structure. According to supervised the observations are separated into two types. First set is input which occurs in the commencement and the next set is output which takes place in the last part of causal chain. Mediating variables can be incorporated in the model between input and output. All the observation are assumed by latent variables. Basically the supervised models leaves the inputs indetermined. If input is accessible then this model is not required. if input values are missing then we cannot finish off anything about output. If input is modeled then the lost input is not a difficulty because it is measured as latent variables which approaches under unsupervised learning[10].

Fig 2 Shows the differences in the casual structure. it is also feasible to have a blend of the both learning because the output observation are understood to be rooted by both latent variables and input observation. When evaluated with supervised learning there is additional opportunities to study better and complex model in unsupervised learning. This is because in supervised learning we are trying to discover the relation between the two sets of observation, so the complication in learning the assignment amplified leads to fewer number of steps between two sets. This is why supervised learning in practice cannot find out deep hierarchies model.

In unsupervised learning it can still even carry on from the launch of observation to the conclusion even more deeply in the level of representation[9]. Each spared steps hierarchy needs to find out only one thing so the learning time enlarges which results in the raise of the level in the model.

In the fundamental relationship between the input and the output the observation is complex and there approaches a big gap and it can be simply bridge using unsupervised learning. This is evidently shown in the fig 3. Instead of finding the relationship between the model we can build a model inside it from upwards of both sets, which results the observation in every level of abstraction and the gap seems to be easier to be filled.

**Supervised Learning**          **Unsupervised Learning**
Observations (inputs)              Latent Variable

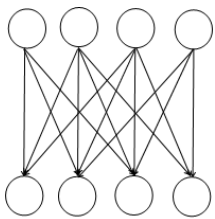   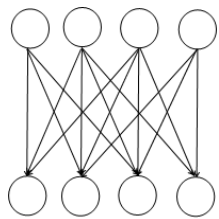

Observations (output)              Observations

Fig 2 . Show the differences in the causal structure in both the learning

Latent Variables

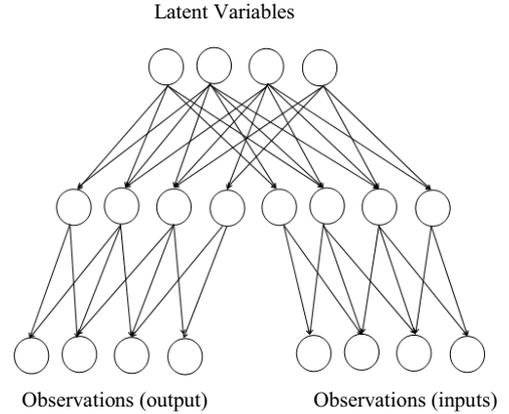

Observations (output)        Observations (inputs)
Fig 3. Represents that the unsupervised learning can be used to fill the gap in-between inputs and outputs.

### III. GROUND RULES

With a small labeled training set we consider a classification problem $\{(x_l^{(1)}, y^{(1)}, x_l^{(2)}, y^{(2)}), ..., (x_l^{(m)}, y^{(m)})\}$. Every input $x_l^{(i)} \in \chi$ is a class label assigned to $y^{(i)} \in y = \{1, 2, ..., k\}$. For the use of non-Gaussian data, the binary vector can be represented as $x_l^{(i)} \in \chi = \{0,1\}^k$ or integer vector $x_l^{(i)} \in \chi = \{0,1,2...\}^k$. For the Deep Framework we should also assume like large set of unlabeled data without subscript $\{x^{(1)}, x^{(2)}, ..., x^{(r)}\}$.

### IV. SPARSE CODING - GAUSSIAN

Generally Gaussian's Sparse uses unlabeled input vector $x \in \Re^k$ to identify basis $b_1, b_2, ..., b_n \in \Re^k$ vector where $x$ is represented as linear combination of basis vector $x \approx \sum_j b_j s_j$. Here $s \in \Re^n$ called activations for the input $x$ [5]. Vector $x$ is originated from Gaussian distribution mean $\eta = \sum_j b_j s_j$ and $\sigma^2 I$ covariance. It follows Laplacian Prior $P(s) \propto \prod_j \exp(-\beta |s_j|)$ some constant $\beta$. Given unlabeled $\{x^1, x^2, ..., x^{(r)}\}$ corresponding activations $\{s^{(i)}, ..., s^{(r)}\}$ and the basic vector $\{b_j\}$ is obtained by

$$\max_{\{b_j\},\{s^{(i)}\}} \prod_i P(x^{(i)} | \{b_j\}, \{s^{(i)}\}) P(s^{(i)})$$

The Optimization Problem

$$\min_{\{b_j\},\{s^{(i)}\}} \frac{1}{2\sigma^2} \sum_i \| x^{(i)} - \sum_{j=1}^n b_j s_j^{(i)} \|^2 + \beta \sum_{i,j} |s_j^{(i)}|$$

Subj.To        $\| b_j \|^2 \leq C, \forall_j = 1, ..., n$

This optimization is Convex for the basic vector and it helps the use of unlabeled data mainly for High - Level Representation[7].

## V. DISCRETE INPUTS FOR SELF TAUGHT LEARNING

To apply the above algorithm to Deep Learning, binary input vector $x \in \{0,1\}^k$. Probabilistic assumption for Gaussian Sparse $P(x|\eta = \sum_j b_j s_j)$. It is a Poor fit to non-Gaussian data. To enforce decomposition $x \approx \sum_j b_j s_j$. Thus it leads to useful basic vector $x \approx \sigma(\sum_j b_j s_j)$, where $\sigma(v) = \left[\frac{1}{1+e^{-v_1}}, \frac{1}{1+e^{-v_2}}, \ldots\right]$. This makes a promise to better form of Sparse Algorithm[4]

**L$_1$ - Regularized Convex Optimization Algorithm**
**Input:** Initialize $s := \vec{0}$.
$B \in \Re^{k \times n}, x \in \Re^k$, Threshold $\in$
**While** objective value decrease at last step $> \in$ **do**
Diagonal Matrix $\Lambda$ compute with $\Lambda_{ii} = a''((Bs)_i)$
Computational Vector $z = \Lambda^{-1}(T(x) - a'(Bs)) + Bs$
Initialize search feature - sign at s, compute
$\hat{s} = \arg\min_{s'} \|\Lambda^{1/2} B s' - \Lambda^{1/2} z\|^2 + \beta \|s'\|_1$
set $s := (1-t)s + t\hat{s}$ where $t$ is established by backtracking line-search to reduce the objective function
**end while**

## VI. DEEP LEARNING APPLICATION

In Deep Learning given $m$ labeled set $\{(x_l^{(1)}, y^{(1)}), (x_l^{(2)}, y^{(2)}), \ldots, (x_l^{(m)}, y^{(m)})\}$. Each $x_l^{(i)} \in \Re^n$, where "$l$" indicates labeled example. Given a set $k$ unlabeled example $x_u^{(1)}, x_u^{(2)}, \ldots, x_u^{(k)} \in \Re^n$

Given labeled and unlabeled set Deep Learning proposes a hypothesis $h : \Re^n \to \{1, \ldots, C\}$

## VII. HIGHER - LEVEL REPRESENTATION

We use a little bit of modified version of Sparse coding algorithm for learning Higher -Level Representations[8].
Given unlabeled data $\{x_u^{(1)}, \ldots, x_u^{(k)}\}$, each $x_u^{(i)} \in \Re^n$.
The optimization problem

$\min_{b,a} \quad \sum_i \|x_u^{(i)} - \sum_j a_j^{(i)} b_j\|_2^2 + \beta \|a^{(i)}\|_1$
Subj. to $\quad \|b_j\|_2 \leq 1, \forall j \in 1, \ldots, s$

The optimization variables $b = \{b_1, b_2, \ldots, b_s\}$ is basis vector each $b_j \in \Re^n$ and $a = \{a^{(1)}, \ldots, a^{(k)}\}$ is activation each $a^{(i)} \in \Re^s$

Here $a_j^{(i)}$ is activation of $b_j$ (basis) for input $x_u^{(i)}$

**Deep Algorithm using Sparse Coding:**

**input** Labeled Set
$T = \{(x_l^{(1)}, y^{(1)}), (x_l^{(2)}, y^{(2)}), \ldots, (x_l^{(m)}, y^{(m)})\}$
Unlabeled data $\{x_u^{(1)}, x_u^{(2)}, \ldots, x_u^{(k)}\}$
**Output** a Learned Classifier for classification task
**Algorithm**: unlabeled data $\{x_u^{(i)}\}$ solve the optimization problem to obtain $b$
compute feature for classification task for new unlabeled training set $\hat{T} = \{(\hat{a}(x_l^{(i)}), y^{(i)})\}_{i=1}^m$
where
$\hat{a}(x_l^{(i)}) = \arg\min_{a^{(i)}} \|x_l^{(i)} - \sum_j a_j^{(i)} b_j\|_2^2 + \beta \|a^{(i)}\|_1$
a classifier $c$ is obtained by applying algorithm, to the unlabeled training set $\hat{T}$
**return** learned classifier c

## VIII. CLASSIFIER : NEURAL NETWORK

Our goal is to use Sparse coding and to find high - Level representation of unlabeled data in deep learning. The task is to find the best classifier for the implementation. This approach is mainly used for Sparse algorithm which is designed for deep learning.

Neural Network Classifier Consists of Neurons i.e Units. The units are arranged in a layer which converts the input vector into output. Each and every unit first takes an input and applies some functions on it and then it passes the input to next level. Networks are said to be feed - forward, which means it feeds the first input's output to the next level which is input to the next level. It does not gives the feedback to the preceding stage so it is entitled as feeding forward. weightings is also applied to the each units signals which passes from one level to another. These weighting actually help or tune the particular problem to adapt the neural network.

Using the Neural Network classifier data is applied in the Matlab. To increase the accuracy in classifier the input latent variables is reduced as vector for preserving correlations which are important for the original data set. By reducing the Dimensionality in the sample data the result seems to be more accurate.

## IX. DISCUSSION

In this paper we presented a general method to use Sparse coding for unsupervised learning with the help of unlabeled data mainly for High - Level Representation. Since Sparse coding does not support unlabeled data or non-Gaussian data we derived an algorithm for the optimization usage which came out as L$_1$-regularized convex optimization algorithm. With the help of this algorithm we derived an Deep Learning algorithm for Higher -Level Representation. Overall our result suggest that $L_1$- regularized convex optimization algorithm can help Higher - Level Representation of documents form unlabeled data and this knowledge is useful in classification problem. We believe this model could be applied generally to other problems, where large amount of unlabeled data are available.

## X. FUTURE WORK

The model represented in the paper consists of two algorithm, which indirectly or directly related the Gaussian Sparse coding with Deep Learning. We have only implemented the model to the images, whereas we like to extend it to Text, Audio and Video related data. It really Seems to be challenging yet this extension when fully related and applied to all the types of multimedia content, It can be applied in Robotics and mainly in their Perception Task.